\documentclass{article}

\usepackage[accepted]{icml2020}
\usepackage[utf8]{inputenc}
\usepackage{tabularx}
\usepackage{latexsym}
\usepackage{url}
\usepackage{multirow}
\usepackage{lipsum}
\usepackage{url}
\usepackage{xspace}
\usepackage{amsmath}
\usepackage[colorlinks=true,allcolors=blue]{hyperref}
\usepackage{amssymb}
\usepackage{mathrsfs}
\usepackage{caption,subcaption}
\usepackage{booktabs}

\usepackage{pgf,tikz,pgfplots}
\pgfplotsset{compat=1.15}
\usetikzlibrary{arrows}
\usepackage{graphicx} 
\usepackage{natbib}
\usepackage{graphicx}

\newcommand{\myparagraph}[1]{\vspace{0.2cm}\noindent\textit{#1}~}

\icmltitlerunning{Chemical Embeddings}

\begin{document}

\twocolumn[
\icmltitle{Word Embeddings for Chemical Patent Natural Language Processing}
\icmlsetsymbol{equal}{*}
\begin{icmlauthorlist}
	\icmlauthor{Camilo Thorne}{to}
	\icmlauthor{Saber Akhondi}{to}
\end{icmlauthorlist}
\icmlaffiliation{to}{Elsevier}
\icmlcorrespondingauthor{Camilo Thorne}{camilo.thorne@gmail.com}
\icmlkeywords{Biomedical NLP, Distributed Representations}
\vskip 0.3in
]

\printAffiliationsAndNotice{} 


\begin{abstract}
We evaluate chemical patent word embeddings against known biomedical embeddings
and show that they outperform the latter extrinsically and intrinsically. We also show that
using contextualized embeddings can induce predictive models of reasonable performance
for this domain over a relatively small gold standard.
\end{abstract}

\section{Introduction}

The chemical industry undoubtedly depends on the discovery of new chemical compounds. However, new chemical compounds are often disclosed first in patent documents \cite{SengerBPG15}, and only months or years later make it to scientific publications.  Thus, most chemical compounds are only immediately available in patent documents \cite{BREGONJE2005309}. As the number of new chemical patent applications has been drastically increasing \cite{MURESAN20111019}, it is now crucial to develop natural language processing (NLP) approaches to automatically extract information from chemical patents \cite{AkhondiBaz001}.

A key tool in this endeavour are word embeddings \cite{baroni-etal-2014-dont}. Embeddings are crucial to derive word, sentence and text-level features in state-of-the-art neural models such as neural named entity recognition (NER) models. Large scale Word2Vec and contextualized embeddings have been developed for e.g., the biomedical \cite{Pyysalo:2013b,jin2019probing} and drug \cite{DBLP:conf/acl-louhi/Segura-BedmarSM15} domains. Smaller embeddings for analytic chemistry such as Mat2Vec \citep{tshitoyan2019unsupervised} (covering materials science) have also been proposed. All such embeddings were trained on scientific papers (PubMed corpus). More recently, \citet{zhai-etal-2019-improving} learnt and successfully applied to chemical named entity recognition large scale Word2Vec and contextualized ELMo embeddings \emph{learnt over chemistry patents} (a 1 billion word corpus), the so-called CheMU embeddings. 

In this paper we address the issue of \emph{evaluating} the quality of these chemistry patent-specific embeddings against its predecessors. Two methods are usual in these kind of comparisons \cite{conf/emnlp/SchnabelLMJ15,journals/jbi/WangLARWSKL18}. On the one hand, \emph{extrinsic} evaluation, in which the impact of each embedding on a prediction task -- chemical NER in this paper -- is reported. On the other hand, \emph{intrinsic} evaluation, where we qualitatively analyze the quality of the distributed (semantic) representations each embedding assigns to chemical words. To this end it is customary to compare the top 10 most similar terms returned by each embedding to a fixed chemical word -- ``ibuprofen", a known anti-inflammatory drug, in this paper -- or similarity query. We show that chemical patent embeddings outperform their predecessors both extrinsically and intrinsically.

\section{Experiments}

\paragraph{Data}

We use for our experiments a small gold standard sampled from two known chemical NER patent corpora, the SCAI corpus~\cite{klinger2008detection} and the Biosemantics corpus~\cite{akhondi2014annotated}. The SCAI corpus focuses on chemicals written using the UIPAC name standard\footnote{See: \url{https://iupac.org}}. In addition, we sampled IUPAC-annotated portions of the Biosemantics corpus (that covers a much wider variety of chemical entity types) to use as validation set. See Table~\ref{tab:SCAI_table}. 

\begin{figure}[tb]
    \centering
    \includegraphics[scale=0.27]{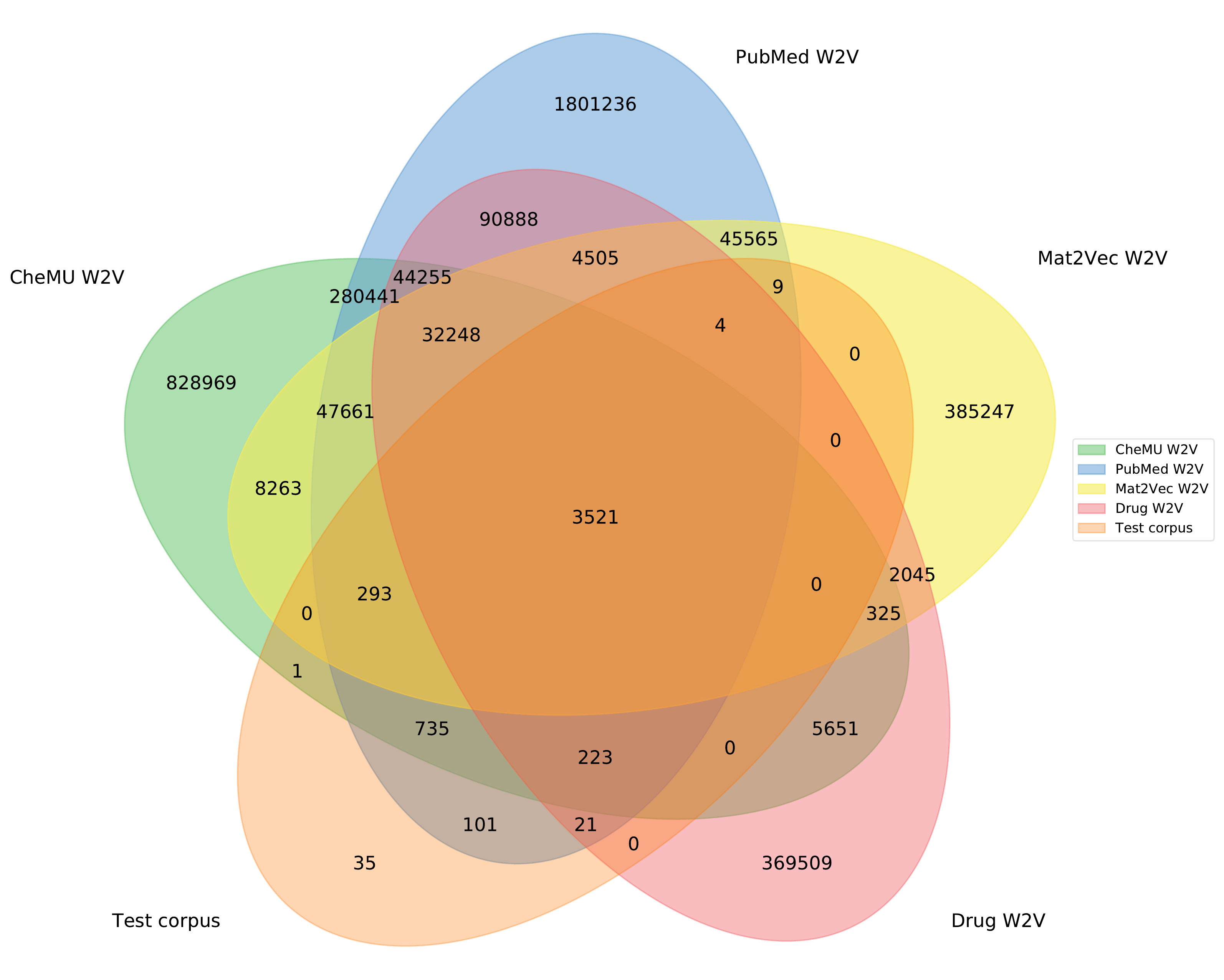}
    \caption{Vocabulary overlaps among Word2Vec embeddings and test corpus. 
    Mat2Vec, PubMed, drug and CheMU embeddings share in total 35,769 words.
    }
    \label{fig:overlaps}
\end{figure}

We observed large vocabulary overlaps among the Word2Vec embeddings, and with the test set (3,521 words), see Figure~\ref{fig:overlaps}. We exploit this fact to qualitatively compare the ELMo and Word2Vec embeddings over the test set vocabulary during the intrinsic evaluation.

\begin{table}[tb]
    \caption{Our training and test sets come from the SCAI corpus; the validation 
    set from the Biosemantics corpus.}
    \centering
    \begin{tabular}{@{}l@{~}l@{~}r@{~~}r@{}}
        \toprule
        Split          & Entities 							& Tokens\\ 
        \midrule
        Train    	& 731 IUPAC, 212 Modifier, 73 Partiupac 	&  33,457 \\
        Validation	& 240 IUPAC 						& 4,654 \\
        Test          	& 48 ~IUPAC, 2 Modifier				& 28,240\\ 
        \bottomrule
    \end{tabular}
    \label{tab:SCAI_table}
\end{table}


\begin{figure}[tb]
    \centering
    \resizebox{.35\textwidth}{!}{
    \newcommand{\binode}[4] {%
    \node (#3) at (-0.6,0) [draw, align=center,minimum width=1cm] {#1};
    \node (#4) at (0.6,0) [draw, align=center,minimum width=1cm] {#2};
}
\begin{tikzpicture}
\tikzset{
 compnode/.style={draw,minimum size=1em,line width=1pt,circle}
}
\pgfsetarrowsend{latex}
\matrix[column sep=2ex, row sep=2ex]{
   & \node (tag0) {O};
   & \node (tag1) {B-IUPAC};
   & \node (tag2) {O};
   \\
   & \node[draw, minimum width=1.1cm] (crf0) {CRF};
   & \node[draw, minimum width=1.1cm] (crf1) {CRF};
   & \node[draw, minimum width=1.1cm] (crf2) {CRF};
   \\
   & \node (c10) {$\oplus$};
   & \node (c11) {$\oplus$};
   & \node (c12) {$\oplus$};
   \\
   & \node[draw, minimum width=1.1cm] (h20) {GRU};
   & \node[draw, minimum width=1.1cm] (h21) {GRU};
   & \node[draw, minimum width=1.1cm] (h22) {GRU};
   \\
   \\
   & \node[draw, minimum width=1.1cm] (h10) {GRU};
   & \node[draw, minimum width=1.1cm] (h11) {GRU};
   & \node[draw, minimum width=1.1cm] (h12) {GRU};
   \\
   & \node (c00) {$\oplus$};
   & \node (c01) {$\oplus$};
   & \node (c02) {$\oplus$};
   \\
   & \binode{word\\emb.}{char.\\emb.}{x00}{x01};
   & \binode{word\\emb.}{char.\\emb.}{x10}{x11};
   & \binode{word\\emb.}{char.\\emb.}{x20}{x21};
  \\
   & \node (i0) {and};
   & \node (i1) {azathioprine};
   & \node (i2) {have};
   \\
 };
 \path
  (i0) edge (x00)
  (i0) edge (x01)
 (c00) edge (h10)
 (h20) edge (c10)
  (c10) edge (crf0)
(crf0) edge (tag0)
  (i1) edge (x10)
  (i1) edge (x11)
 (c01) edge (h11)
 (h21) edge (c11)
  (c11) edge (crf1)
(crf1) edge (tag1)
  (i2) edge (x20)
  (i2) edge (x21)
 (c02) edge (h12)
 (h22) edge (c12)
  (c12) edge (crf2)
(crf2) edge (tag2)
 (h11) edge (h12)
 (h10) edge (h11)
 (h22) edge (h21)
 (h21) edge (h20)
;
\path[-]
 (crf0) edge (crf1)
 (crf1) edge (crf2)
;
 \draw (x01) -- (c00);
 \draw (x00) -- (c00);
 \draw (x11) -- (c01);
 \draw (x10) -- (c01);
 \draw (x21) -- (c02);
 \draw (x20) -- (c02);
 \path[bend left=60]
 (h10) edge (c10)
 (c00) edge (h20)
 (h11) edge (c11)
 (c01) edge (h21)
 (h12) edge (c12)
 (c02) edge (h22)
;
\end{tikzpicture}
    }
    \caption{Architecture of the (neural) NER system used.}
    \label{fig:model}
\end{figure}
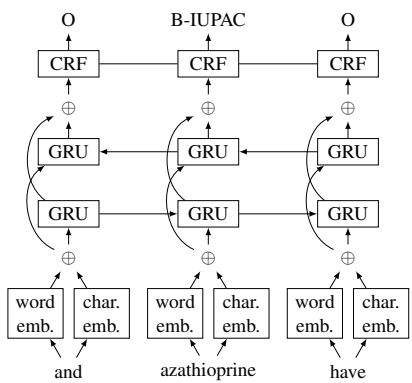

\begin{table}[tb]
    \caption{Overview of the embeddings studied in this paper.}
    \centering
    \begin{tabular}{@{}rrr@{}}
        \toprule
        Embedding               	& Words 		& Dimensions\\
        \midrule
        Mat2Vec W2V           &    529,686      & 200 \\
        PubMed W2V           	&    2,351,706   & 200   \\
        Drug W2V                 	&     553,195     & 420   \\
        CheMU W2V          	&    1,252,586   & 200  \\
        PubMed ELMo          & ---~~~~~ 	&1,204 \\
        CheMU ELMo           & ---~~~~~ 	& 1,204 \\
        \bottomrule
    \end{tabular}
    \label{tab:emb_over}
\end{table}

\paragraph{Extrinsic Evaluation}

For NER -- viz., \emph{extrinsic} evaluation -- we use simplified variants of \citet{DBLP:journals/corr/LampleBSKD16}'s model by using GRU instead of LSTM (stacked) layers, to reduce parameters and prevent overfitting given the small size of our corpus. This is sufficient as we wanted to (a) test if the embeddings could induce reasonable (though not state of the art) performance, and (b) test if different embeddings give rise to different performances. We encoded words using the pre-trained chemical embeddings and a trainable character-level GRU encoder. See Figure~\ref{fig:model}. The models were trained for 50 epochs with early stopping (patience of 5 epochs), 80-dimensional bidirectional GRU character encodings with 0.25 dropout, and 300-dimensional bi-directional GRU token encoder with 0.5 dropout. We used a batch size of 16. As training algorithm, Adam was used, with a learning or decay rate of 0.01 and L2-regularization. We used AllenNLP as our main implementation\footnote{\url{https://allennlp.org/}}. The models were trained on a Tesla T4 GPU with 16GB of GPU RAM.

As Table~\ref{tab:NER_results} shows, using large the scale PubMed Word2Vec embeddings increases only marginally F1 score w.r.t.\ Mat2Vec (baseline model). They are both beaten by a wide margin by the drug-specific Word2Vec embeddings. Using patent-specific chemical Word2Vec embeddings also yields the best F1 score for Word2Vec embeddings. The best results are obtained with the ELMo embeddings, which outperform the Word2Vec embeddings again by a wide margin, and interestingly allow the model to achieve a reasonable F1 score of  72,41\%. It is also interesting to observe that PubMed ELMo embeddings contain sufficient domain knowledge as to reach a close (and also reasonable) F1 score of 70.15\%.

\begin{table}[tb]
    \caption{Impact of the different chemical embeddings on chemical NER (sorted by F1 score).}    
    \centering
    \begin{tabular}{l@{\qquad}c@{\qquad}lc@{\qquad}}
        \toprule
        Word Embedding  & F1 & $\Delta$ (F1)\\
        \midrule
        Mat2Vec W2V  & 26.89\% & ~~~~---\\
        PubMed W2V   & 27.23\% & +~\,0.3\%\\
        Drug W2V        & 48.48\% & +21.3\%\\
        CheMU W2V   & 53.24\% & +~\,4.8\%\\
        PubMed ELMo& 70.15\% & +16.9\%\\
        CheMU  ELMo & 72.41\% & +~\,2.3\%\\
        \bottomrule
    \end{tabular}
    \label{tab:NER_results}
\end{table}

\begin{table*}[tb]
    \caption{Top 10 similarity lists (``ibuprofen" query).}
    \centering
    \begin{tabular}{@{}cccccc@{}}
    \toprule
    CheMU   & PubMed & CheMU & PubMed & Drug & Mat2Vec \\ 
    ELMo    & ELMo   & W2V   & W2V    & W2V  & W2V     \\ 
    \midrule
    tacrine	    & atropine	    & aspirin           &	aspirin	    & pronounced	& drug \\
    ondansetron & ondansetron   & clopidogrel       & ondansetron   & ultrastructure& drugs \\
    aspirin	    & sulfamethoxazole	& prednisolone	& clopidogrel	& mimics	    & aspirin\\
    clopidogrel	& aspirin	    & azathioprine	    & propranolol	& surgical	    & sulfamethoxazole\\
    dipyridamole& tacrine	    & atropine	        & placebo	    & favorable	    & propranolol\\
    atropine	& trimethoprim	& nifedipine	    & tacrine	    & intestine	    & trimethoprim\\
    prednisolone& propranolol	& sulfamethoxazole	& nifedipine	& trained       & norfloxacin\\
    propranolol	& prednisolone	& dipyridamole	    & prednisolone	& extinct	    & estradiol	\\
    trimethoprim& clopidogrel	& propranolol	    & mg	        & slightly	    & antibiotics \\
    nifedipine	& papaverine	& papaverine	    & topical	    & combination	& nifedipine \\
    \bottomrule
    \end{tabular}
    \label{tab:top10_embedding_ibu}
\end{table*}


\begin{table}[tb]
    \caption{Overlap of similarity lists (``ibuprofen" query).}
    \centering
    \begin{tabular}{@{}l@{~}c@{~}c@{~}c@{~}c@{~}c@{}}
    \toprule
& PubMed & Drug & Mat2Vec & CheMU & PubMed\\
& W2V & W2V & W2V & ELMo & ELMo\\
\midrule
W2V CheMU 	& 0.33 & 0.25 	& 0.25 	& 0.43 	& 0.54\\ 
W2V PubMed 	& ---  & 0.25 	& 0.18 	& 0.43 	& 0.54\\ 
W2V Drug  	& --- 	& --- 		& 0.18 	& 0.25 	& 0.18\\ 
W2V Mat2Vec 	& --- 	& --- 		& --- 		& 0.11 	& 0.33\\ 
ELMo CheMU 	& --- 	& --- 		& --- 		& --- 		& 0.54\\ 
    \bottomrule
    \end{tabular}
    \label{tab:top10_sim_i}
\end{table}


\begin{figure}[tb]
	\centering
	\includegraphics[scale=0.3]{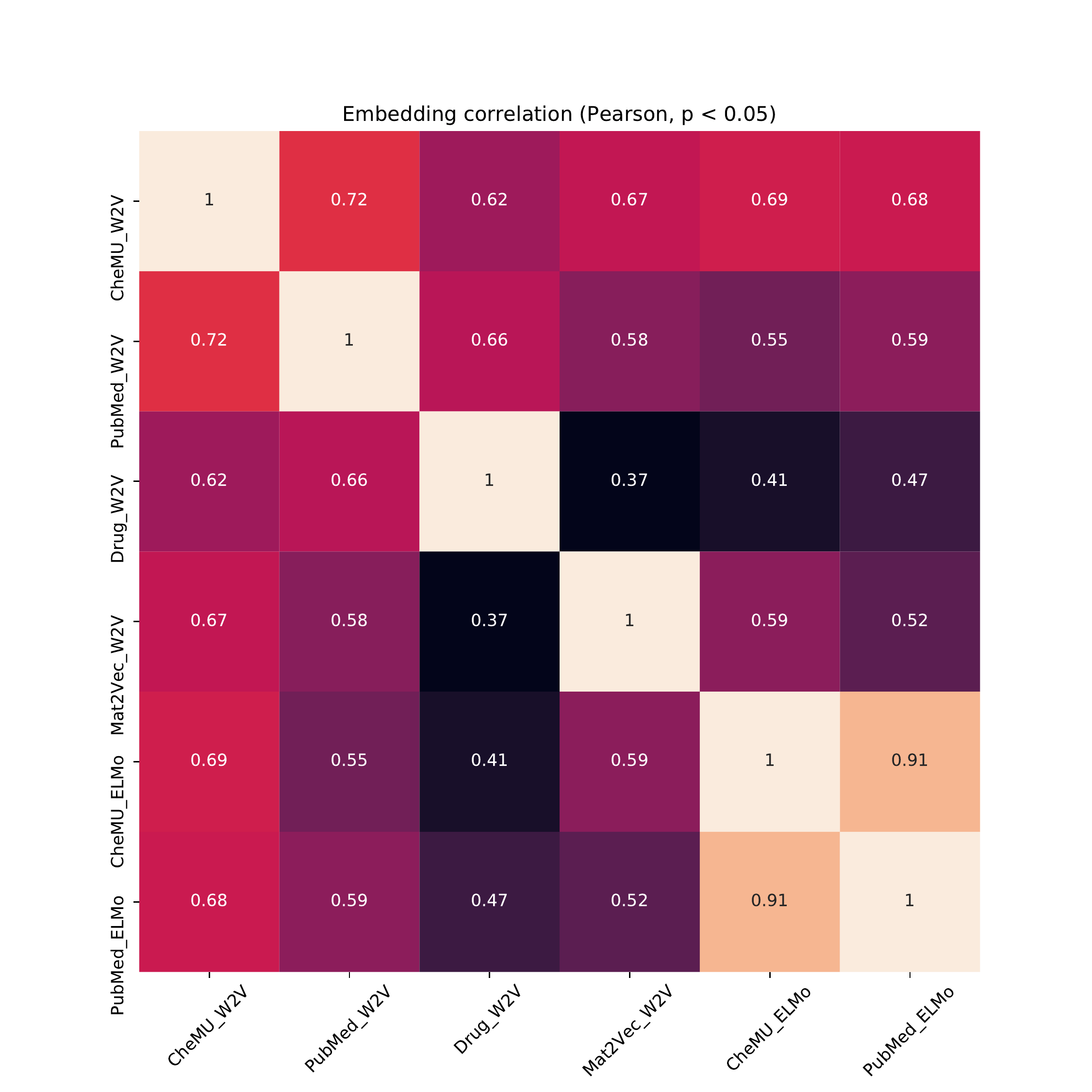}
	\caption{Embedding correlation (test corpus vocabulary).}
	\label{fig:correlation}
\end{figure}

\paragraph{Intrinsic Evaluation}

For qualitative or \emph{intrinsic} evaluation, we adopted the following strategy to generate comparable embeddings. We restricted the vocabulary of Word2Vec embeddings to the vocabulary of our test corpus (we excluded stop and function words). To compare these restricted Word2Vec embeddings to the ELMo embeddings, we computed an ELMo embedding for each occurrence of a test corpus content word, and then averaged all such embeddings to derive \emph{ad hoc} single corpus-level Word2Vec embeddings. While such approach potentially discards contextual information, it can still preserve syntactic and semantic word type information, given that contextualised embeddings tend to assign similar embeddings to words that assume similar grammatical and semantic roles within a sentence. The dimensionality of the embeddings was also standardized, by reducing, using singular value decomposition (SVD), to 200 dimensions the 420-, and 1,024-dimensional embeddings. We carried out three analysis.

\myparagraph{Similarity analysis.}
We chose a drug entity, ``ibuprofen", mentioned in the test corpus and retrieved its topmost 10 most similar words using cosine -- $\textit{sim}(w,w') = (\vec{w} * \vec{w}') / || \vec{w} \cdot \vec{w}'||$ -- similarity. Ideally, since ibuprofen is a drug, what we expect to see in such similarity lists are names of drugs or chemical compounds.

The results obtained align -- with some caveats -- with the results observed over the NER model. As Table~\ref{tab:top10_embedding_ibu} shows, chemical patent embeddings produce better rankings than their more generic or non-patent specific counterparts. Furthermore, ELMo embeddings again show better results. Mat2Vec, Drug and PubMed word embeddings return common nouns (``drugs''), abbreviations (``mg"), adjectives ("topical") or verbs (``mimics"), whereas chemical patent Word2Vec and ELMo embeddings return only drugs or chemicals. Interestingly as well, they return as their topmost most similar term names of substances (``aspirin", ``tacrine", ``atropine'') with somewhat similar anti-inflammatory properties.

\myparagraph{Agreement analysis.}
These trends are largely confirmed when we check how much these lists semantically align. As chemical terms tend to be ambiguous, we used DrugBank\footnote{DrugBank is a database of chemical substances, see: \url{https://www.drugbank.ca}} to normalise the terms $w$ into their InChI $i(w)$ chemical identifiers\footnote{An InChI identifier defines a unique representation of a molecule, see: \url{https://www.inchi-trust.org}.}. Finally, we measured the set similarity -- $\textit{sim}(W,W') = | i(W) \cap i(W)' | / | i(W) \cup i(W)'|$ -- of the ensuing normalised lists to measure how much the different embeddings ``agree" on their understanding of ``ibuprofen".

As Table~\ref{tab:top10_sim_i} shows, chemical patent W2V and ELMo embeddings substantially align (0.43 and 0.54 set similarity resp.). On the other hand, the drug and Mat2Vec embeddings do not seem to align well to any other embedding. Interestingly, PubMed Word2Vec embeddings exhibit a higher than expected alignment with chemical patent embeddings.

\myparagraph{Correlation analysis.}
Finally, we measured the degree of correlation between the embeddings. As Figure~\ref{fig:correlation} shows, both ELMo embeddings correlate very highly (0.91)\footnote{Which may explain their similar properties.}. Chemical patent and PubMed Word2Vec embeddings moderately correlate with all embeddings, whereas Mat2Vec and drug embeddings give rise to lower correlation scores.

\section{Conclusions}

We have studied the quality of embeddings trained over chemical patents against those of embeddings of close biomedical domains. Our experiments show that patent specific embeddings outperform extrinsically other embeddings by giving rise to better F1 scores in chemical NER (72.41\% on our test corpus). Correlation and similarity analysis indicate that they also outperform them intrinsically, and provide a better understanding of the chemistry domain. They also show that contextualized (ELMo) embeddings yield globally better results, and that generic but large scale PubMed ELMo embeddings (that cover the full life sciences domain) yield reasonable results for the chemistry domain.

\paragraph{Aknowledgments.}
We thank Karin Verspoor, Christian Druckenbrodt and Zenan Zhai for the discussions that resulted in the early versions of this paper.

\bibliographystyle{icml2020}
\bibliography{references}

\newpage
\onecolumn

\section*{Appendix}

We display here two additional sets of visualizations. On the one hand, a visualization of the embeddings compared, and on the other hand, a molecular similarity ranking
for Table~\ref{tab:top10_embedding_ibu}. To generate the (chemical) structural diagrams and measure molecular fingerprint similarity
we used RDKit\footnote{\url{https://www.rdkit.org/}}. We ignored terms in the lists that could not be normalized. For the embedding visualizations, we picked a random set of
100 content words from the SCAI test corpus (covered by all the embeddings). In parallel to this, we applied the t-SNE algorithm to reduce the input 200-dimensional embeddings
to 2 dimensions.

The visualizations generally align with what we observed in the body of the paper. Fingerprint similarity rankings align more with the ELMo embedding rankings. Likewise, the visualizations of the ELMo-derived embeddings tend to converge, whereas all the others diverge considerably.


\begin{figure}[h]
    \centering
    \includegraphics[width=0.9\textwidth]{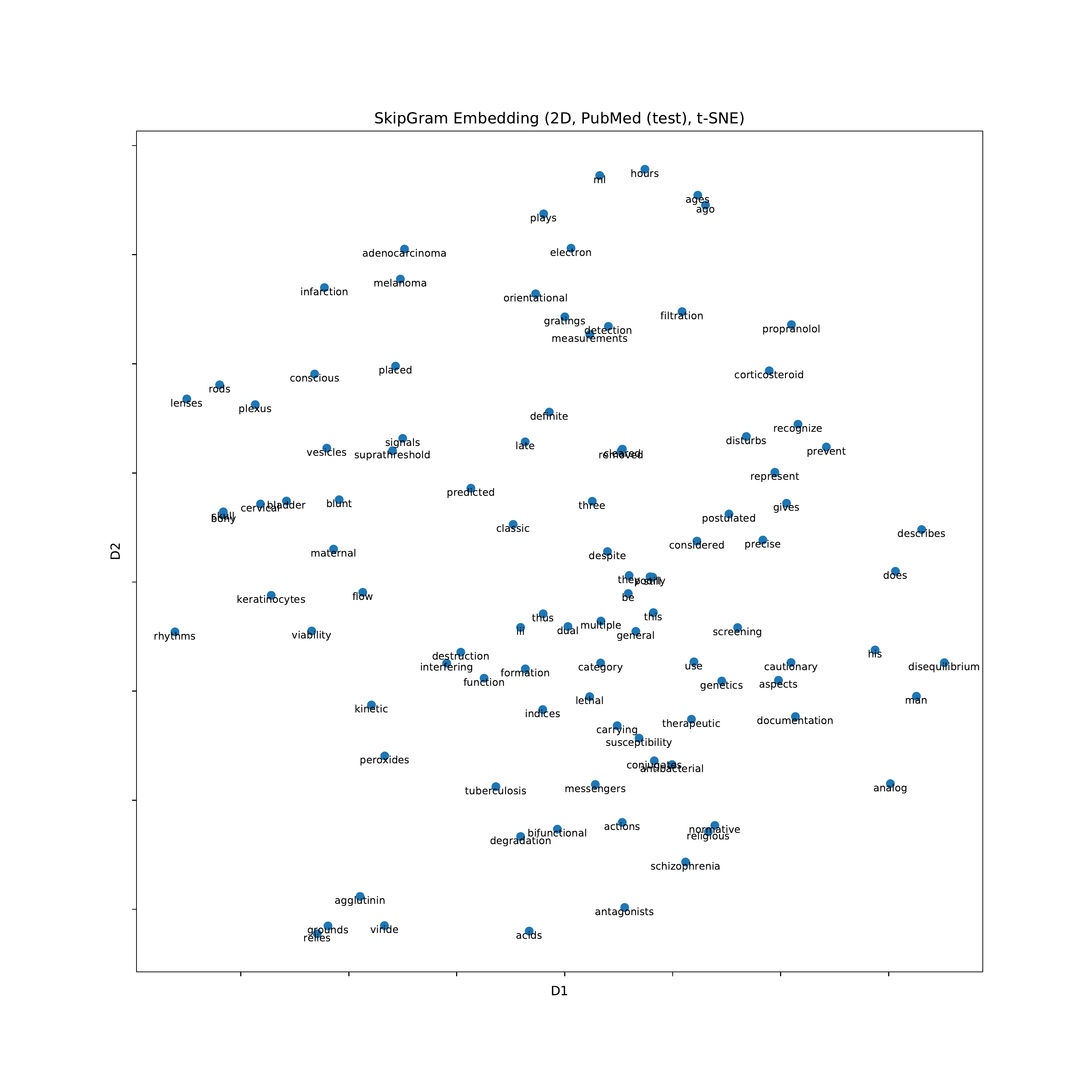}
    \caption{PubMed W2V visualization.}
\end{figure}    
\begin{figure}[h]
    \centering
    \includegraphics[width=0.9\textwidth]{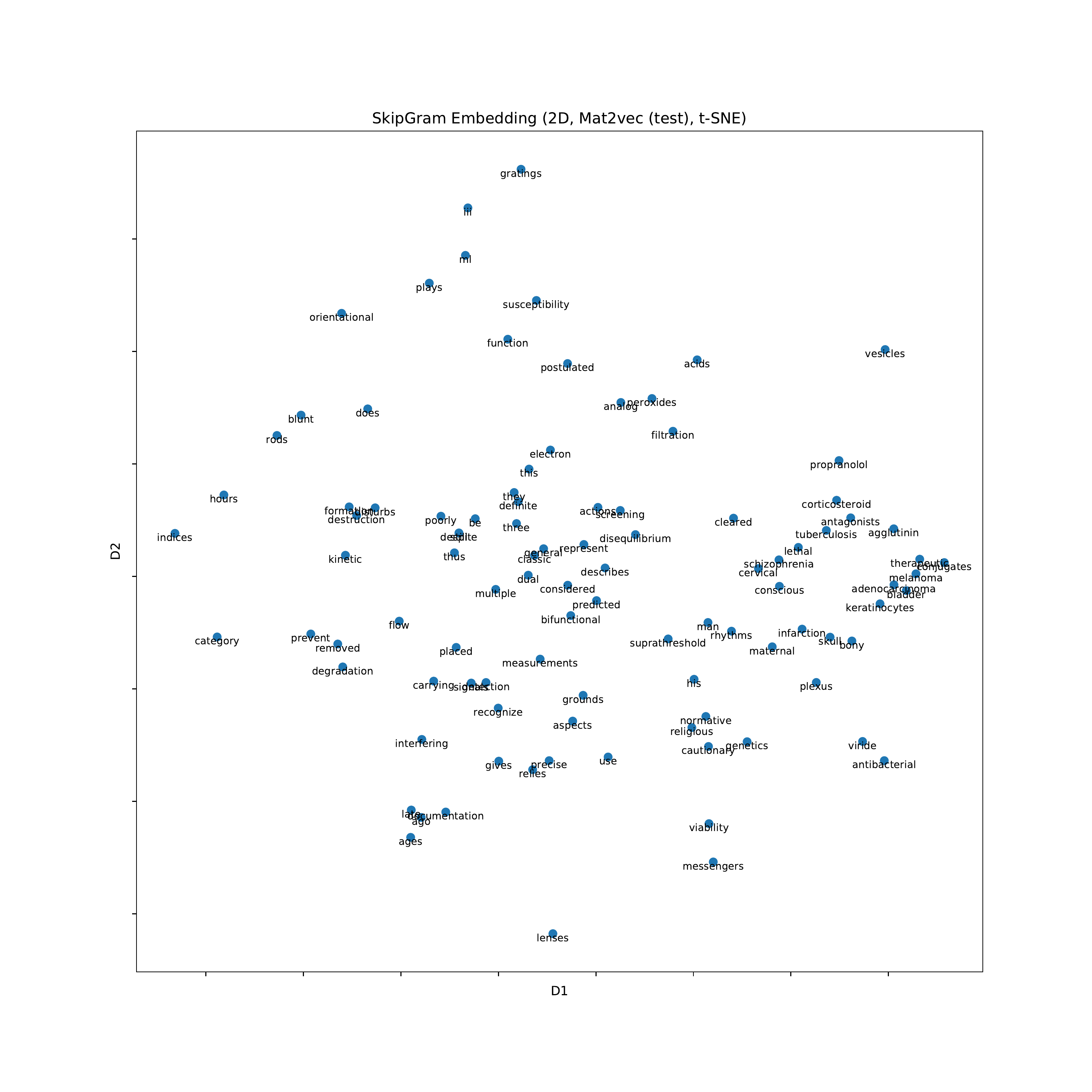} 
    \caption{Mat2Vec W2V visualization.}
\end{figure}  
\begin{figure}[h]
    \centering    
    \includegraphics[width=0.9\textwidth]{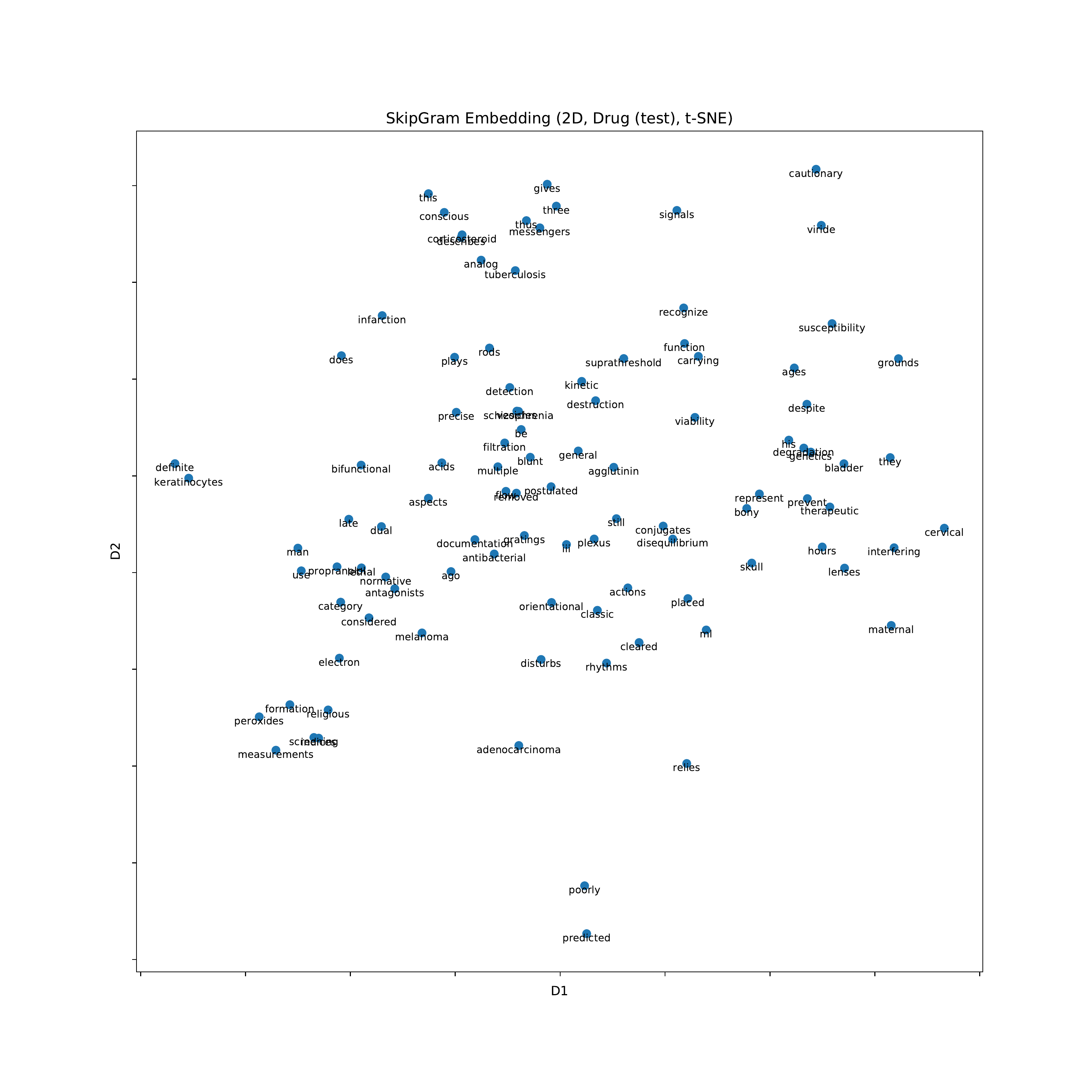}
    \caption{Drug W2V visualization.}
\end{figure}  
\begin{figure}[h]
    \centering    
    \includegraphics[width=0.9\textwidth]{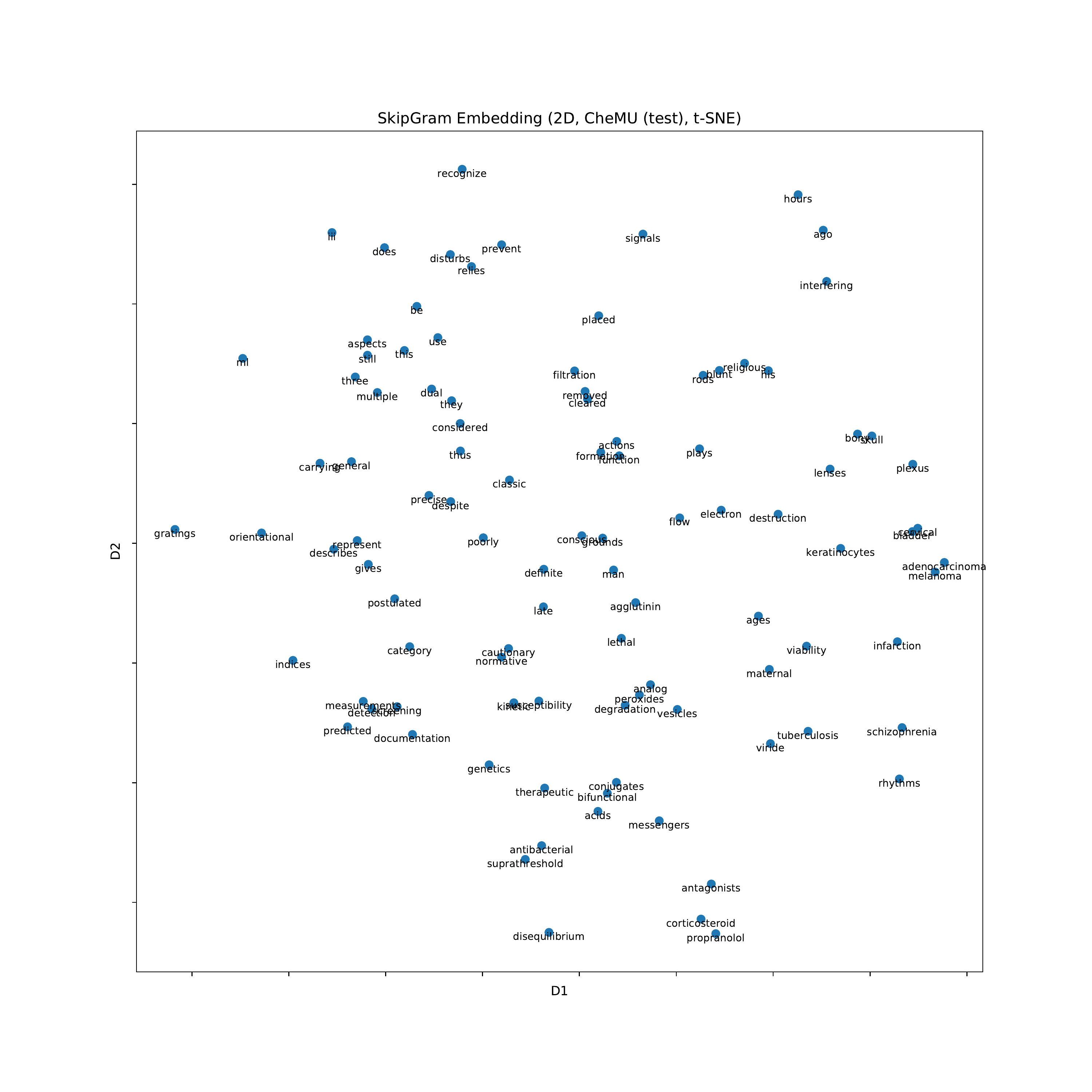} 
    \caption{CheMU W2V visualization.}
\end{figure}  
\begin{figure}[h]
    \centering    
    \includegraphics[width=0.9\textwidth]{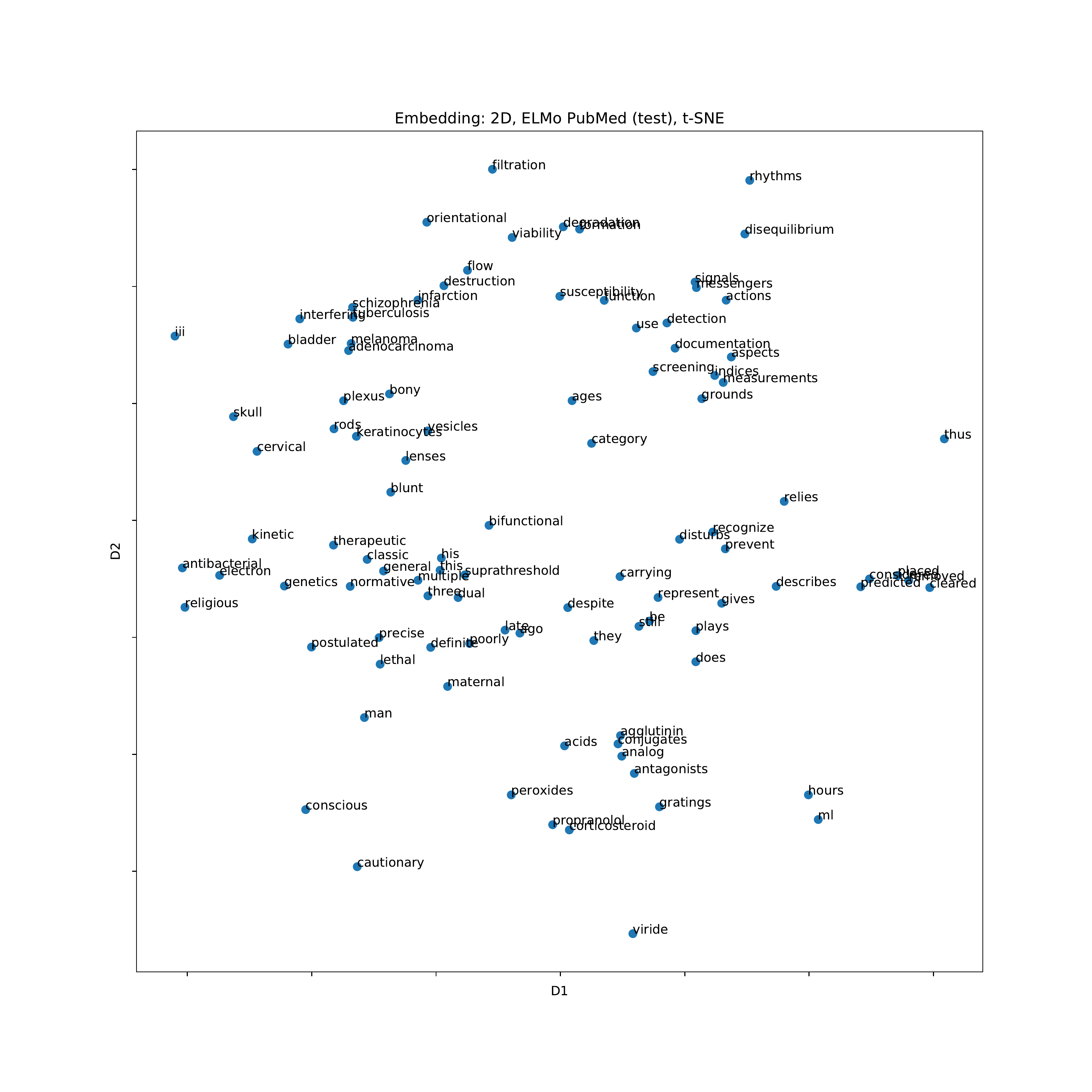}
    \caption{PubMed ELMo visualization.}
\end{figure}  
\begin{figure}[h]
    \centering    
    \includegraphics[width=0.9\textwidth]{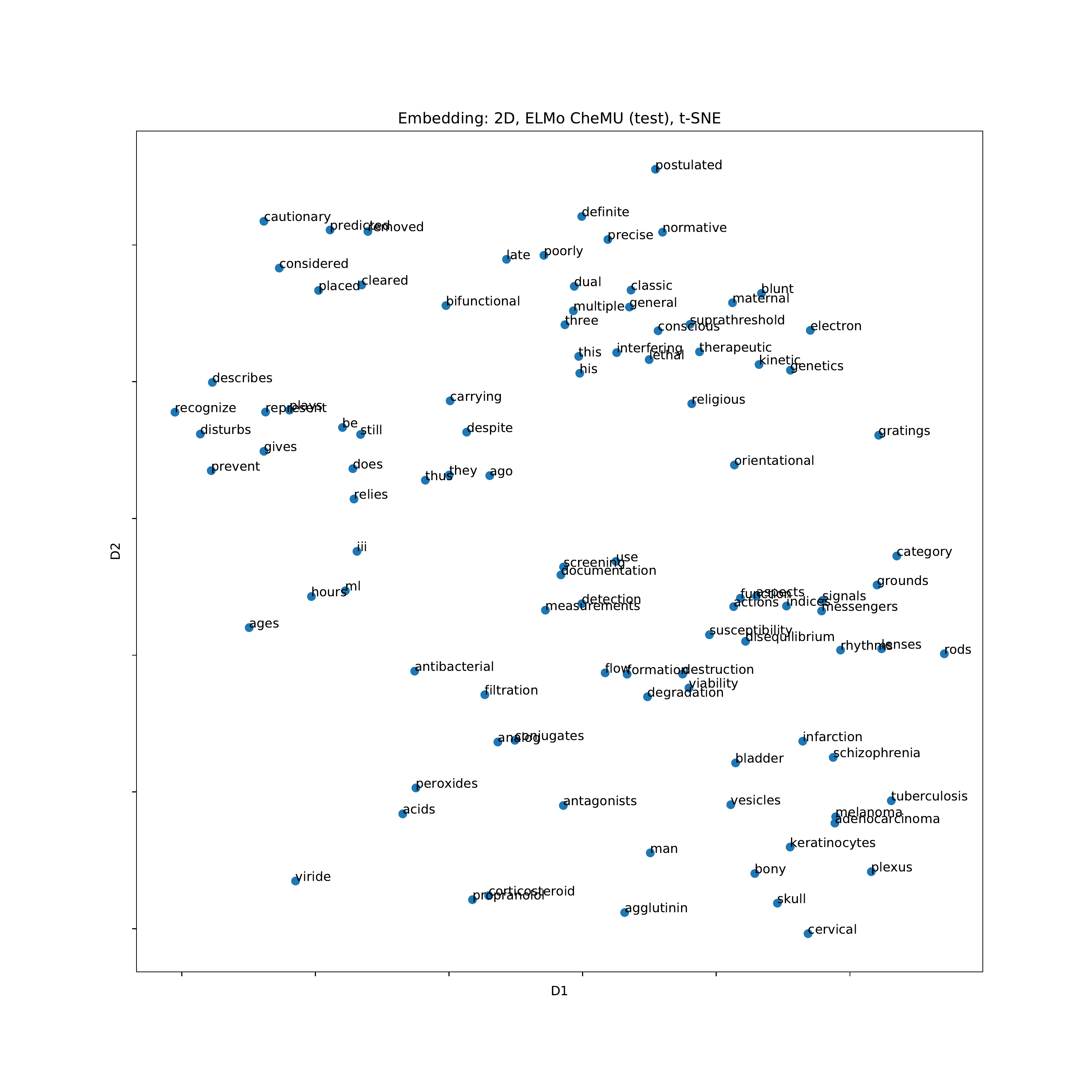}
    \caption{CheMU ELMo visualization.}
\end{figure}  

 \begin{figure}[tb]
    \centering
    \includegraphics[width=0.8\textwidth]{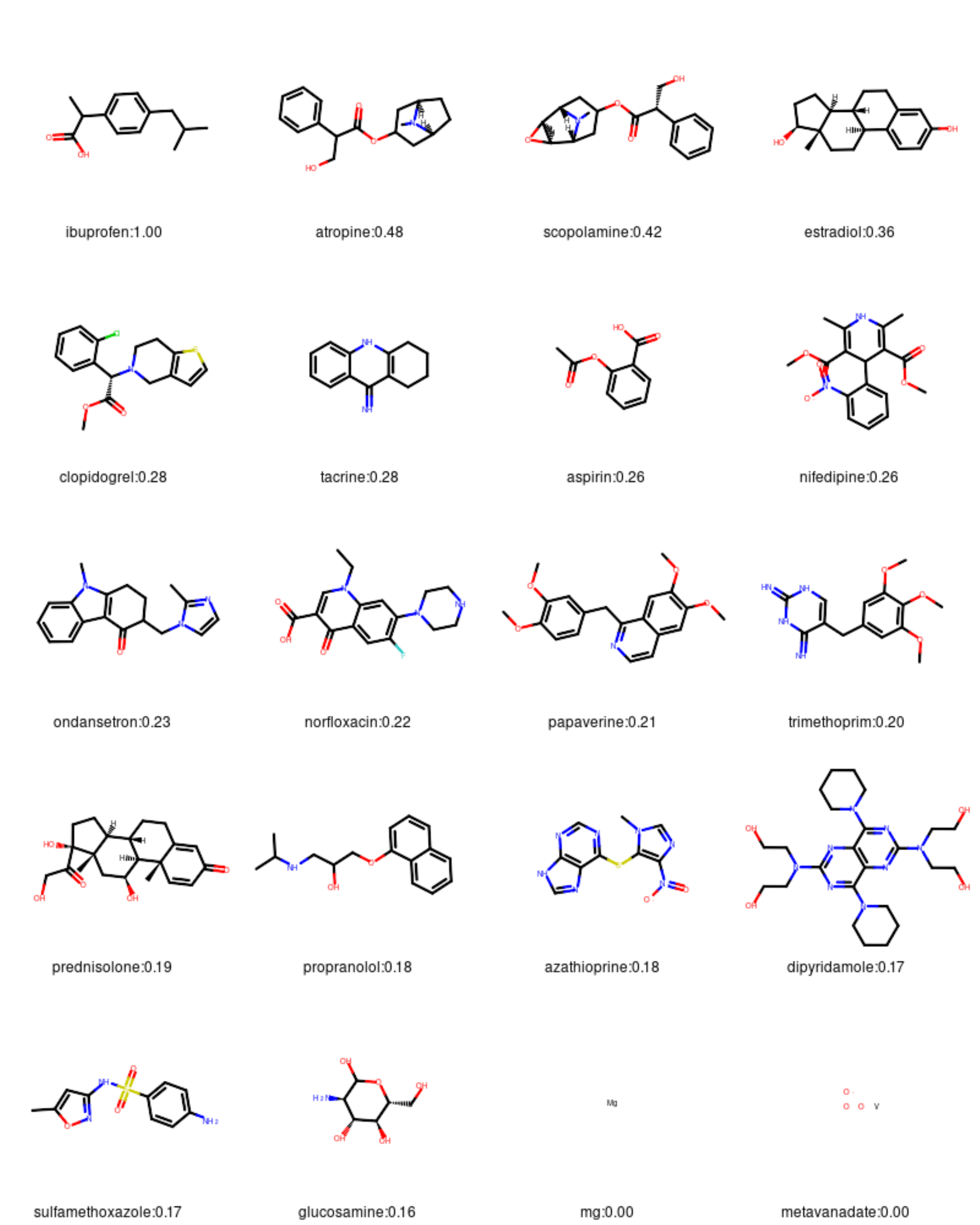}
    \caption{Chemical structures of the (chemically normalizable) terms  returned by the ``ibuprofen" query, sorted by molecular (fingerprint) similarity.}
    \label{fig:structs_ibu}
\end{figure}


\end{document}